\documentclass[letterpaper]{article}
\usepackage[utf8]{inputenc}
\usepackage[english]{babel}
\usepackage[T1]{fontenc}
\usepackage{geometry}
\usepackage{flairs}
\usepackage{times}
\usepackage{helvet}
\usepackage{courier}
\usepackage{subfig}
\usepackage{graphicx}
\usepackage{tabularx}
\usepackage{longtable}
\usepackage{bm}
\usepackage{mathtools} 
\usepackage{booktabs}
\usepackage{makecell}
\usepackage{multirow}
\usepackage{multicol}
\usepackage{diagbox}
\usepackage{amsfonts,bm,amsmath,tikz,url}
\usetikzlibrary{matrix}

\usetikzlibrary{arrows,plotmarks,decorations.markings,trees,shapes}
\tikzstyle{n}=[ellipse,draw=black!100,fill=black!10,line width=.7pt,minimum width=1cm,align=center,minimum height=.5cm]
\tikzstyle{nodo}=[ellipse,draw=black!100,fill=black!30,line width=.7pt,minimum width=1.2cm,text width=2.2cm,align=center,minimum height=.7cm]
\tikzstyle{Qnodo}=[ellipse,draw=black!100,fill=black!10,line width=.7pt,minimum width=1.2cm,,text width=2.2cm,align=center,minimum height=.7cm]
\tikzstyle{arco}=[draw=black!80,line width=1pt, postaction={decorate}, decoration={markings,mark=at position 1.0 with {\arrow[ draw=black!80,line width=.7pt]{>}}}]
\tikzstyle{decision} = [rectangle, draw, fill=black!100,text=white, text width=4.5em, text badly centered, node distance=3cm, minimum height=3em]
\tikzstyle{block} = [rectangle, draw, fill=blue!20, text width=5em, text centered, rounded corners, minimum height=3em]
\tikzstyle{line} = [draw, -latex']
\tikzstyle{cloud} = [draw, ellipse,fill=red!20, node distance=3cm, minimum height=2em]
\begin{document}
\title{Intelligent Tutoring Systems by Bayesian Nets with Noisy Gates\thanks{We thank the Swiss National Science Foundation (Grant n.187246) for financial support.}}
\author{Alessandro Antonucci, Francesca Mangili, Claudio Bonesana, Giorgia Adorni\\
Istituto Dalle Molle di Studi sull'Intelligenza Artificiale (IDSIA), Lugano - Switzerland\\
{\tt \{alessandro,francesca,claudio,giorgia.adorni\}@idsia.ch}}
\maketitle
\begin{abstract}
\begin{quote}
Directed graphical models such as Bayesian nets are often used to implement intelligent tutoring systems able to interact in real-time with learners in a purely automatic way. When coping with such models, keeping a bound on the number of parameters might be important for multiple reasons. First, as these models are typically based on expert knowledge, a huge number of parameters to elicit might discourage practitioners from adopting them. Moreover, the number of model parameters affects the complexity of the inferences, while a fast computation of the queries is needed for real-time feedback. We advocate logical gates with uncertainty for a compact parametrization of the conditional probability tables in the underlying Bayesian net used by tutoring systems. We discuss the semantics of the model parameters to elicit and the assumptions required to apply such approach in this domain. We also derive a dedicated inference scheme to speed up computations. 
\end{quote}
\end{abstract}
\section{Introduction}
Developing \emph{intelligent tutoring systems} (ITSs) able to support learners without the supervision of a human teacher and provide feedback and customized instructions has been a classical field of application for AI since its early stages. Unsurprisingly, the change of paradigm from rule-based AI systems performing logical reasoning to (probabilistic) uncertain reasoning systems also impacted the ITS design \cite{mayo2001bayesian}. In more recent times the growing need of e-learning technologies brought a renovated interest on ITSs \cite{guo2021evolution}. While ITS technologies are benefiting from the more recent advances of deep learning to obtain a more flexible human-machine interaction, using probabilistic approaches in the assessment and tutoring part remains essential. 

A typical modeling approach for ITSs includes an assessment step in which the competence profile of the learner is described by a set of latent variables (to be called \emph{skills} in the following) whose actual value affects some manifest variables such as the answers given to different questions or the reactions to feedback. This is the case of \emph{item response theory} (IRT), the most popular approach for such models. Yet, when coping with multiple skills, IRT might be ineffective in describing skill correlations and more expressive probabilistic formalisms might be needed instead. Among them, probabilistic models based on directed graphs such as Bayesian nets (BNs) represents a highly interpretable option, often adopted for ITS implementation  \cite{millan2000adaptive}. 

\begin{figure}[htp!]
\centering
\begin{tikzpicture}
\node[nodo] (s1)  at (0,0) {\tiny Finger Mult. ($X_1$)};
\node[nodo] (s2)  at (4,0) {\tiny Long Mult. ($X_2$)};
\node[Qnodo] (q1)  at (0.,-1) {\tiny $3 \times 4=?$ ($Y_1$)};
\node[Qnodo] (q2)  at (4,-1) {\tiny $13 \times 14=?$ ($Y_2$)};
\draw[arco] (s1) -- (q1);
\draw[arco] (s2) -- (q1);
\draw[arco] (s2) -- (q2);
\end{tikzpicture}
\caption{An ITS to test multiplicative skills based on a BN.}
\label{fig:bn}
\end{figure}
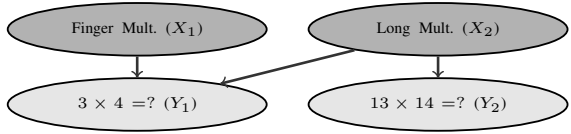

This allows to graphically depict the relations between skills and questions as in the example in Fig. \ref{fig:bn}. Although BN arc orientation should not necessarily reflect a causal interpretation, in practice, graphs implementing a ITS often have a bipartite structure with arcs from the skills to the questions but not vice versa. As a consequence, each question receives incoming arcs from the skills that are relevant to answer the question. The corresponding elicitation involves a number of parameters exponential with respect to the number of relevant skills. This might be a serious issue discouraging ITS practitioners from using these tools in their applications because of a too demanding elicitation process when many skills are affecting the answer to a question. Another serious issue is related to inferential complexity. ITS decisions are based on an inference in the underlying BN, a notorious NP-hard task. Complexity depends on the graph topology being in practice exponential in the graph treewidth. Again, questions affected by many skills prevent a fast computation of inference leading to a feedback for the learner.

A solution to the above issues is to reduce the number of parameters in the local models of the BN by deterministic functions corresponding to degenerate probabilistic parameters. Yet, this is based on rigid assumptions, while higher realism is achieved by adding noise to such deterministic relations. The most prominent example of this kind is the so-called \emph{noisy-OR gate} \cite{pearl1988probabilistic}. This allows to reduce the number of parameters to elicit from exponential to linear.
Similar advantages also concern the inference.

We advocate the use of noisy gates for the elicitation of the parameters of a ITS based on a BN. When coping with ITSs a disjunctive relation over the skills might be unrealistic and we consider the use of more general logical functions (e.g., conjunctive). We also clarify the semantic of the parameters in such models and their impact on the corresponding inferences by deriving a number of monotonicity results. 
Overall, we obtain a compact and transparent model elicitation allowing for real-time ITS feedback.

\section{Bayesian Nets}
We use uppercase letters to denote variables, lowercase for states and calligraphic for the whole set of states. Thus, for instance, variable $X$ takes its generic value $x$ from set $\mathcal{X}$. In the Boolean case $\mathcal{X}=\{0,1\}$ and notation $\neg x$ refers to the the negation of state $x$. A probability mass function (PMF) $P(X)$ over $X$ is a non-negative map $\mathcal{X}\to \mathbb{R}$ such that $\sum_{x\in\mathcal{X}} P(x)=1$. Given a joint PMF $P(X_1,X_2)$, we can compute conditional probabilities by Bayes rule, e.g., $P(x_1|x_2) = P(x_1,x_2)/P(x_2)$, where $P(x_2)$ is obtained by marginalization, i.e., $P(x_2)=\sum_{x_1\in\mathcal{X}_1} P(x_1,x_2)$. Note that the conditional probability is defined only if $P(x_2)>0$. A (conditional) PMFs is denoted as $P(X_1|x_2)$. Similarly, a conditional probability table (CPT) is a collection of conditional PMFs, e.g., $P(X_1|X_2)=\{P(X_1|x_2)\}_{x_2\in\mathcal{X}_2}$.

Let $\mathcal{G}$ denote a directed acyclic graph whose nodes are in one-to-one correspondence with the variables in $\bm{X}:=(X_1,\ldots,X_N)$. A Bayesian net (BN) consists in a collection of CPTs $\{P(X|\mathrm{Pa}(X))\}_{X\in\bm{X}}$ where $\mathrm{Pa}(X)$ are the \emph{parents}, i.e., the immediate predecessors of $X$ according to $\mathcal{G}$. A BN defines a joint PMF $P(\bm{X})$ such that
$P(\bm{x}) = \prod_{i=1}^N P(x_i|\mathrm{pa}(X_i))$, for each $\bm{x} \in \bm{\mathcal{X}}$, where the values of $x_i$ and $\mathrm{pa}(X_i)$ are those consistent with $\bm{x}$ for each $i=1,\ldots,N$ \cite{koller2009probabilistic}. Such compact factorization is based on the independence relations induced from $\mathcal{G}$ by the \emph{Markov condition}, i.e., each node is assumed to be independent of its non-descendants non-parents given its parents. BN inference consists in the computation of queries based on the joint PMF. In particular we are interested in \emph{updating} tasks consisting in the computation of the (posterior) PMF for a (single) variable $X_q \in \bm{X}$ given some evidential information about other variables $\bm{X}_E \subseteq \bm{X}$, that are assumed to be in a state $\bm{x}_E$. This corresponds to:
\begin{equation}\label{eq:updating}
P(x_q|\bm{x}_E) = \frac{\sum_{\bm{x}'} \prod_{i=1:N} P(x_i|\mathrm{pa}(X_i))}{\sum_{\bm{x}',x_q} \prod_{i=1:N} P(x_i|\mathrm{pa}(X_i))}\,,
\end{equation}
where $\bm{X}':=\bm{X}\setminus \{\bm{X}_E,X_q\}$ and, here and in the following, the domains of a sum is left implicit for the sake of readability. The task in Eq. \eqref{eq:updating} is NP-hard in the general case \cite{koller2009probabilistic}. Exact inference schemes such as message propagation allow to compute exact inference for simple (e.g., singly-connected) topologies of $\mathcal{G}$ in polynomial time. In the general case, exact inference is exponential with respect to the \emph{treewidth} of $\mathcal{G}$, while models with high treewidth should be updated by approximate schemes. The situation is therefore especially critical for models with high maximum \emph{indegree} (i.e., number of parents) as this involves both large CPTs and treewidth.
\section{Noisy Gates}
We focus on Boolean models. The extension to non-Boolean (but ordinal) variables is discussed in the conclusions. 

\paragraph{Disjunctive Gates.} We consider the quantification of the CPT for a variable $Y$ in a BN. Say that $(X_1,\ldots,X_n)$ are the parents of $Y$. The elicitation of $P(Y|X_1,\ldots,X_n)$ requires a number of parameters exponential in the number of parents $n$ (namely $2^n$ as we cope with Boolean variables). A \emph{noisy-OR gate} \cite{pearl1988probabilistic} requires instead the elicitation of $n$ parameters only, say $\{\lambda_i\}_{i=1}^n$, with, for each $i=1,\ldots,n$, $\lambda_i\in [0,1]$. The corresponding CPT is specified as follows:
\begin{equation}\label{eq:noisy}
P(Y=0|x_1,\ldots,x_n) = \prod_{i=1}^n (\mathbb{I}_{x_i=0}+\lambda_i\mathbb{I}_{x_i=1})\,,
\end{equation}
where $\mathbb{I}_A$ is an indicator function returning one if $A$ is true and zero otherwise.

\begin{figure}[htp!]
\centering
\begin{tikzpicture}
\node[n] (xx1)  at (0,.8) {\tiny $X_1$};
\node[n] (xx2)  at (1.5,.8) {\tiny $X_2$};
\node[] (xx3)  at (3,.8) {\tiny $\ldots$};
\node[n] (xx4)  at (4.5,.8) {\tiny $X_n$};
\node[n] (x1)  at (0,0) {\tiny $X_1'$};
\node[n] (x2)  at (1.5,0) {\tiny $X_2'$};
\node[] (x3)  at (3,0) {\tiny $\ldots$};
\node[n] (x4)  at (4.5,0) {\tiny $X_n'$};
\node[n] (y)  at (2.25,-0.7) {\tiny $Y$};
\draw[arco] (xx1) -- (x1);
\draw[arco] (xx2) -- (x2);
\draw[arco] (xx4) -- (x4);
\draw[arco] (x1) -- (y);
\draw[arco] (x2) -- (y);
\draw[arco] (x3) -- (y);
\draw[arco] (x4) -- (y);
\end{tikzpicture}
\caption{A noisy gate (explicit formulation).}
\label{fig:bn2}
\end{figure}
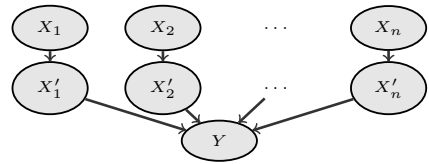

To better understand the semantic of such models, let us introduce $n$ auxiliary Boolean variables $(X_1',\ldots,X_n')$. Consider a BN fragment as in Figure \ref{fig:bn2} where $Y$ is the common child of these variables being in fact their disjunction, i.e., $Y := \bigvee_{i=1}^n X_i'$. Such deterministic relation induces a degenerate (i.e., involving only probabilities equal to zero and one) quantification of the corresponding CPT. We further set the input variable $X_i$ as the unique parent of $X_i'$ for each $i=1,\ldots,n$. The quantification of the (two-by-two) CPT $P(X_i'|X_i)$ is achieved by setting $P(X_i'=0|X_i=0)=1$ and $P(X_i'=0|X_i=1)=\lambda_i$. Accordingly, we can regard the auxiliary variable $X_i'$ as a \emph{inhibition} of $X_i$: if the input $X_i$ is false then $X_i'$ is certainly false, while if $X_i'$ is true then $X_i$ can be false with probability $\lambda_i$. It is straightforward to prove that such BN fragment returns the probabilities in Eq. \eqref{eq:noisy} when the auxiliary variables are summed out. In fact $P(Y=0|x_1,\ldots,x_n)$ can be expressed as:
\begin{equation}\label{eq:noisy3}
\sum_{(x_1',\ldots,x_n')} P(Y=0|x_1',\ldots,x_n') \prod_{i=1}^n P(x_i'|x_i) 
\prod_{i=1}^n P(X_i'=0|x_i)
\,,
\end{equation}
and hence as $\prod_{i=1}^n P(X_i'=0|x_i)$ because the CPT of $Y$ implements a disjunction and the only configuration giving non-zero probability to $Y=0$ is the one where all the parents are in their false state. Eq. \eqref{eq:noisy} trivially follows from Eq. \eqref{eq:noisy3} by the definition of the CPT $P(X_i'|X_i)$.

Accordingly the noisy-OR can be intended as a model of a causal relation between the input variables $(X_1,\ldots,X_n)$ and $Y$. The mechanism is disjunctive and a single true cause is sufficient to make the effect $Y$ true. Yet, a noisy process acting independently and with different probabilities for each input/cause might make the active cause ineffective. 

\paragraph{Conjunctive Gates} The noisy-OR model, as depicted in Figure \ref{fig:bn2}, can be generalized. First, we might replace the disjunction defining the CPT $P(Y|X_1',\ldots,X_n')$ with a different deterministic function $f$. Regarding the CPT $P(X_i'|X_i)$ we keep the assumption 
that one of its two PMFs is deterministic, while the other depends on $\lambda_i$. In particular we want that for $\lambda_i=0$ the CPT implements the deterministic constraint $X_i' \equiv X_i$, while for $\lambda_i=1$ the two PMFs are equal and the CPT expresses the irrelevance of $X_i$ to $X_i'$. This is achieved by setting 
$P(X_i'=\hat{x}|X_i=\hat{x})=1$ and 
$P(X_i'=\hat{x}|X_i=\neg \hat{x})=\lambda_i$. 
We call $\hat{x}$ \emph{distinguished} state. 

In the above general setup, the noisy-OR gate corresponds to set $f$ as a disjunction and $\hat{x}=0$. Setting $f$ as a conjunction and $\hat{x}=1$ defines instead a \emph{noisy-AND}. In the latter case $\lambda_i$ can be intended as an \emph{activation} probability (i.e., if $X_i$ is true also $X_i'$ is certainly true, while if $X_i$ is false it might become true with that probability). The analogous of Eq. \eqref{eq:noisy} for the noisy-AND gate is:
\begin{equation}\label{eq:noisy22}
P(Y=1|x_1,\ldots,x_n) =\prod_{i=1}^n [\mathbb{I}_{x_i=1}+\lambda_i\mathbb{I}_{x_i=0}]\,.
\end{equation}
The derivation is analogous to that in Eq. \eqref{eq:noisy3}, with  non-zero contributions involving only states such that $X_i'=1$. A generalization of Eq. \eqref{eq:noisy} that holds both for the noisy-OR and the noisy-AND gates based on the notion of distinguished state is therefore:
\begin{equation}\label{eq:noisy2}
P(Y=\hat{x}|x_1,\ldots,x_n) =\prod_{i=1}^n [\mathbb{I}_{x_i=\hat{x}}+\lambda_i\mathbb{I}_{x_i=\neg\hat{x}}]\,.
\end{equation}

A further extension can be achieved by making the gate \emph{leaky}, i.e., adding a Boolean variable $X_0'$ that is a parent of $Y$ and setting $P(X_0'=\neg \hat{x})=\lambda_0$. A leaky model is equivalent to a standard noisy gate with an additional parent $X_0$ and corresponding parameter $\lambda_0$, with $X_0$ instantiated in its non-distinguished state. This allows to obtain a leaky version of Eq. \eqref{eq:noisy2} by simply adding on the right-hand side an additional term involving $\lambda_0$ and such that $X_0=\neg \hat{x}$. 

\section{Exploiting noisy gates in BN-based ITS}
Noisy gates allow for a compact specification of the CPTs in a BN. For a compact model specification we adopt a simple structure in the graph underlying the BN: the skills are parentless nodes $X_1,\dots,X_n$ and $X_i=1$ means that the learner possesses skill $i$, the questions are childless nodes $Y_1,\dots,Y_m$ whose parents are the skills relevant to answer that particular question (see, e.g., the BN in Figure \ref{fig:bn}) and $Y_j=1$ describes a correct answer. At the elicitation level, the domain expert should asses the prior PMFs $P(X_i)$ for the different skills and, if we use noisy gates for the questions, $n\times m$ parameters $\lambda_{i,j}$, one for each parent/skill $i$ of each question $j$. In the previous section, the BN in Fig. \ref{fig:bn2} has been shown to represent both a noisy-OR and a noisy-AND provided that $Y_j$ is either the conjunction or the disjunction of all auxiliary nodes $X'_{i,j}$, and that, given $\hat{x}$, then $P(X'_{i,j}=\hat{x}|X_i=\hat{x})=1$ and $P(X'_{i,j}=\hat{x}|X_i=\neg\hat{x})=\lambda_{i,j}$. A disjunction with $\hat{x}=0$ implements a noisy OR, a conjunction with $\hat{x}=1$ a noisy AND. 

\paragraph{Interchangeable Skills (Or).} For disjunctive gates, $Y_j$ changes to the non-distinguished state corresponding to a correct answer if at least one of the auxiliary variables is the non-distinguished state $X'_{i,j}=1$. This model can be used to describe a situation where a single skill is sufficient to answer a specific question (e.g., in Fig. \ref{fig:bn}, multiplications are solved either by fingers or by long multiplication). 
If the learner has the skill $i$, the probability of answering $j$ correctly by means of that skill depends on the question and is equal to $1-\lambda_{i,j}$. E.g., in Fig. \ref{fig:bn}, $1-\lambda_{1,1}$ describes the probability that counting with fingers allows to solve the first multiplication. The fact that the second multiplication is too hard to be solved by fingers is graphically represented by the absence of a direct arc between $X_1$ and $Y_2$. In the noisy-OR model, this translates in the constraint $1-\lambda_{1,2} = 0$. In general, setting $1-\lambda_{i,j} = 0$, implies that having skill $X_i$ has no effects on the capability of the learner to solve question $Y_j$ and that the inability to answer question $Y_j$ does not change the prior opinion about the learner possessing skill $X_i$. On the other hand, $1-\lambda_{i,j} = 1$ means that having skill $X_i$ ensures that the learner correctly answers question $Y_j$. 

In a noisy-OR gate, for a learner missing all skills, all auxiliary variables are in the false state and, therefore, the learner is expected to be always wrong. 
This can be easily solved by making the gate leaky. The probability $P(X'_0=1|X_0=1)=1-\lambda_0$ describes the chances of a random guess; for instance, in a multiple choice question with four options, one of which is correct, one should set the probability $1-\lambda_0=1/4$. 

\paragraph{Complementary Skills (And).} In disjunctive gates, all auxiliary variables must be in their distinguished state $X_{i,j}'=1$ for the answer to be correct. Therefore, a noisy-AND is useful to describe complementary skills, which must be mastered at the same time in order to answer properly. 

In the noisy-AND model, possessing skill $X_i$ implies the ability to apply it to any question $Y_j$, whereas a weakness in skill $X_i$ alone will compromise the success in task $Y_j$ with probability $1-\lambda_{i,j}$. When $1-\lambda_{i,j} = 0$, it follows that lacking $X_i$ in no way can compromise the success in task $Y_j$. 
On the other hand, when $1-\lambda_{i,j} = 1$, task $Y_j$ is such that the lack of skill $X_i$ alone causes the student to fail. 
 
In a noisy AND,  if a learner has all skills, she/he can certainly solve any task, thus excluding the possibility of doing wrong by other reasons than the lack of one of the modeled skills. Again, this can be easily solved by a leaky gate where $P(X_0'=0|X_0=0)=1-\lambda_0$ is the probability of failing the task despite the presence of all skills. 

In both cases, $1-\lambda_{i,j}$ represents the importance of skill $X_i$ with respect to question $Y_j$, being the probability that the non distinguished state of the $i$-th skill will alone lead question $Y_j$ to be in the non distinguished state as well. The extreme cases where  $1-\lambda_{i,j}=0$ and $1-\lambda_{i,j}=1$ represent, respectively, a situation where the skill and the question nodes are conditionally independent, that is, skill $X_i$ is not relevant to question $Y_j$, and a situation where the non-distinguished state of the skill $X_i$ alone implies the non-distinguished state of question $Y_j$. 
\section{Posterior Probabilities}
Here we derive analytical expressions for the inferences of interest in ITSs based on BNs. We focus on the posterior probabilities for the skill of interest updated after the gathering of the answers from the learners. We initially consider a BN as in Fig. \ref{fig:bn2} and derive expressions for the probabilities of a particular skill, say $X_k$, given that the answer is the distinguished state (i.e., $P(X_k|Y=\hat{x})$) or the negation of the distinguished state (i.e., $P(X_k|Y=\neg\hat{x})$). In this way we obtain expressions valid for both noisy-OR and noisy-AND gates. We also characterize the dependence of these expressions with respect to the corresponding parameter $\lambda_k$ and their relation with the marginal PMF $P(X_k)$. 

\paragraph{The answer is the distinguished state.} In a conjunctive (resp. disjunctive) model, $Y$ is true (resp. false) if and only if all its parents $(X_1',\ldots,X_n')$ are in their true (resp. false) states. In other words $Y=\hat{x}$ implies $X_i'=\hat{x}$ for each $i=1,\ldots,n$. Such induced evidence allows in particular to drop the arc connecting $X'_k$ with $Y_k$. By Markov condition:
\begin{equation}\label{eq:or0}
P(X_k|Y=\hat{x})=P(X_k|X_k'=\hat{x})\,.
\end{equation}
The posterior queries of $X_k$ can be therefore obtained by local computations involving only the elements of the CPT $P(X_k|X_k')$ and the marginal $\pi_k:=P(X_k=\neg \hat{x})$. In fact, by Bayes rule, we can obtain from Eq. \eqref{eq:or0} that:
\begin{equation}\label{eq:or01}
P(X_k=\neg \hat{x}|Y=\hat{x}) 
= \frac{\pi_k \lambda_k}{\pi_k \lambda_k + (1-\pi_k)}
\,.
\end{equation}
Eq. \eqref{eq:or01} is a monotonically increasing function of $\lambda_k$, that is always greater or equal than $\pi_k$. 

\paragraph{The answer is the non distinguished state.} If $Y=\neg \hat{x}$, it is not possible to directly propagate the evidence as in the previous case. To compute $P(X_k=\neg \hat{x}|Y=\neg \hat{x}) = P(X_k=\neg \hat{x},Y=\neg \hat{x})/P(Y=\neg \hat{x})$ first notice that:
\begin{equation} \label{eq:trick}
P(X_k=\neg \hat{x},Y=\neg \hat{x})=\pi_k-P(X_k=\neg\hat{x},Y=\hat{x})\,.
\end{equation}

In principle the computation of $P(Y=\neg\hat{x})$ requires a sum over all the $2^n$ joint states of the parents of $Y$. In practice this can be achieved in linear time by a recursion \cite{pearl1988probabilistic}. In fact, by the total probability theorem:
\begin{equation}\label{eq:tot}
P(Y=\hat{x})=\sum_{(x_1,\ldots,x_n)} P(Y=\hat{x}|x_1,\ldots,x_n) \prod_{i=1}^n P(x_i)\,.
\end{equation}
Eq. \eqref{eq:noisy2} allows to rewrite the r.h.s. of Eq. \eqref{eq:tot} as
\begin{equation}\label{eq:fn}
\sum_{(x_1,\ldots,x_n)} \prod_{i=1}^n \left[ (1-\pi_i) \mathbb{I}_{x_i=\hat{x}} + \lambda_i \pi_i \mathbb{I}_{x_i=\neg \hat{x}} \right]\,,
\end{equation}
and, finally, $P(Y=\hat{x}) = \prod_{i=1}^n (1 - \pi_i (1-\lambda_i))$.
Thus, by Eqs. \eqref{eq:or01} and \eqref{eq:trick}, $P(X_k=\neg \hat{x}|Y=\neg \hat{x})$ becomes:
\begin{equation}\label{eq:post2}
\frac{\pi_k-
\pi_k \lambda_k
\prod_{i\neq k} (1 - \pi_i (1-\lambda_i))
}{1-\prod_{j=1}^n (1 - \pi_j (1-\lambda_j))}\,.
\end{equation}
Concerning monotonicity with respect to $\lambda_k$, it can be shown that the derivative of \eqref{eq:post2} is always negative, and therefore $P(X_k=\neg \hat{x}|Y=\neg \hat{x})$ is a monotonically decreasing function of $\lambda_k$, being also greater than or equal to $\pi_k$.

\paragraph{Summary and general case} Eqs. \eqref{eq:or01} and \eqref{eq:post2} allow for an analytical characterization of the effects of an answer on ITSs based on BNs with noisy gates. The monotonicity results with respect to $\lambda_k$ are consistent with an elicitation strategy that intends $1-\lambda_k$ as the relative importance of the $k$-th skill to properly answer question $Y$. The relation with $\pi_k$ implies that, for both noisy-OR and noisy-AND gates, inequality 
$P(X_k=1|Y=0) \leq P(X_k=1) \leq P(X_k=1|Y=1)$, i.e., the qualitative behavior we expect in ITSs when the learner provides wrong or right answers.

Regarding complexity, our formulae can be computed in linear time. Yet, this refers to the updating of a single answer/evidence, that is equivalent to cope with a BN as in Fig. \ref{fig:bn2}. Coping with multiple evidences does not make any difference if the answers are the distinguished states of their gates. In those cases, by Eq. \eqref{eq:or0}, we can move the evidence to the auxiliary variables of the gate and disconnect them from the question. In practice, the evidence can be embedded in the model by re-defining the marginal PMF of $X_k$ with the posterior values in Eq. \eqref{eq:or01} separately for each evidence and each $k$. As a result, we might focus on the updating of the non-distinguished answers by considering a BN where only the corresponding gates are present. 
\section{Conclusions}
We advocate BNs with noisy (conjunctive or disjunctive) gates for the implementation of ITSs. A natural outlook is the extension to ordinal variables. The has been already proposed in \cite{diez1993parameter}, but providing an analogous characterization requires a dedicated effort. Following the ideas in \cite{antonucci2021a}, it seems possible to extend the procedure derived in this paper from Bayesian to \emph{credal nets}, i.e., BNs with interval parameters. The model in \cite{antonucci2011c} might be considered for that.
\bibliographystyle{flairs}
\bibliography{biblio}
\end{document}